# Physics of Language Models: Part 2.2, How to Learn From Mistakes on Grade-School Math Problems

Tian Ye
tye2@andrew.cmu.edu
CMU / Meta FAIR

Zicheng Xu
zichengxu@meta.com
Meta FAIR

Yuanzhi Li
Yuanzhi.Li@mbzuai.ac.ae
MBZUAI

Zeyuan Allen-Zhu
zeyuanallenzhu@meta.com
Meta FAIR

August 28, 2024

(version 1)*

**Abstract**

Language models have demonstrated remarkable performance in solving reasoning tasks; however, even the strongest models still occasionally make reasoning mistakes. Recently, there has been active research aimed at improving reasoning accuracy, particularly by using pretrained language models to "self-correct" their mistakes via multi-round prompting. In this paper, we follow this line of work but focus on understanding the usefulness of incorporating "error-correction" data directly into the pretraining stage. This data consists of erroneous solution steps immediately followed by their corrections. Using a synthetic math dataset, we show promising results: this type of pretrain data can help language models achieve higher reasoning accuracy directly (i.e., through simple auto-regression, without multi-round prompting) compared to pretraining on the same amount of error-free data. We also delve into many details, such as (1) how this approach differs from beam search, (2) how such data can be prepared, (3) whether masking is needed on the erroneous tokens, (4) the amount of error required, (5) whether such data can be deferred to the fine-tuning stage, and many others.

---

*Project page + video: https://physics.allen-zhu.com/part-2-grade-school-math/part-2-2.
We would like to thank Lin Xiao, Beidi Chen, Chunting Zhou for many helpful conversations. We would like to extend special thanks to Lucca Bertoncini, Liao Hu, Caleb Ho, Wil Johnson, Apostolos Kokolis, Parth Malani, Alexander Miller, Junjie Qian and Shubho Sengupta from Meta FAIR; Henry Estela, Rizwan Hashmi, Lucas Noah, and Maxwell Taylor from the Meta Cloud Foundation team; as well as Ian Clark, Gourab De, Anmol Mann, and Max Pfeifer from W&B. Without their invaluable support, the experiments in this paper would not have been possible.

# 1 Introduction

Language models have achieved near-human-level performance in various tasks, including math solving, coding, and natural language understanding [1, 2, 18, 21, 31]. However, their problem-solving skills are still imperfect, sometimes resulting in logical errors. Recently, there have been numerous attempts to improve the reasoning accuracy of language models.

One promising approach is to use a verifier to check the correctness of the language model's output [9, 12, 23, 28, 30]. Interestingly, some studies show that language models can "self-verify" [15, 27]: They can be prompted to verify the correctness of their own generation, thereby improving overall accuracy. An illustrative example is shown in Figure 1.

This leads to the following fundamental questions:

*If a language model can correct its own mistakes after generation, (1) Why does it make those mistakes to begin with? (2) Why doesn't it correct the mistakes immediately during generation, instead of waiting until after?*

There are many works that attempt to understand question (1). Notably, studies such as [6, 17, 22] have shown that many mistakes arise due to "distribution shift" — the training distribution of the language model differs from the prompts used during testing. Thus, even if the training data is error-free, language models can still make mistakes during generation.

Much less work focuses on question (2). While correcting mistakes after generation is a valid approach to improve a language model's accuracy, it is more desirable to correct mistakes immediately as they occur, **such as "$A \Rightarrow B$, oh I made a mistake, actually $A \Rightarrow C$."** Doing so during generation can save inference tokens (the model does not need to continue generating based on an erroneous step) and simplify the inference framework (eliminating the need to call the model multiple times to verify and correct the answer), see Figure 1. Thus, can we train language models to directly perform (2)?

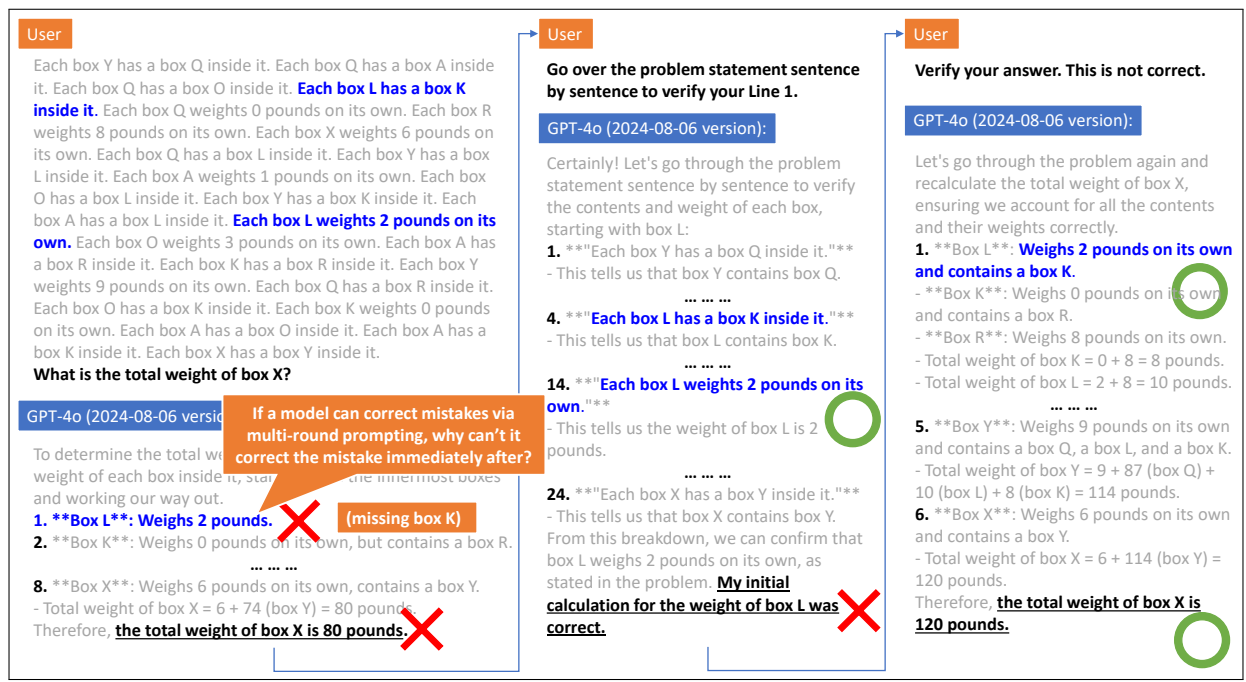

Figure 1: An illustration of how GPT-4o self-verifies and corrects its own mistakes.
    **Observation:** Correcting mistakes during generation, rather than after, can save inference tokens (the model avoids generating useless tokens following an erroneous step) and simplify the inference process (eliminating the need for multiple rounds of verification and correction).

A natural conjecture for why current language models cannot perform (2) is an alignment issue: There might be little to no training data in the language model's corpus to teach it to correct errors immediately. After all, people are unlikely to write an erroneous sentence followed by immediate correction in internet articles.[1] However, **even if** we are given enough such "**retry**" training data (i.e., error + immediate correction such as "$A \Rightarrow B$, oh I made a mistake, actually $A \Rightarrow C$"), **is it clear that the language model can improve its overall reasoning correctness?**

- **Potential harm of next-token prediction on data with mistakes:** Training on data with mistakes can be dangerous. By minimizing the next-token prediction objective, the model might learn to generate mistakes (such as "$A \Rightarrow B$") instead of writing the correct steps (i.e., $A \Rightarrow C$). One could try to mitigate this by masking out the erroneous steps. However, unlike the "error correction after entire generation" type of training data where we can simply mask out the entire generation,[2] here, the error and correction steps are intertwined, making it much harder to mask out the error only.

- **Unclear benefits over training on "perfectly error-free" data:** Intuitively, if our final goal is to have model output correct reasoning steps to improve the reasoning accuracy, then why would training on "$A \Rightarrow B$, oh I made a mistake, actually $A \Rightarrow C$" be better than training directly on the correct step "$A \Rightarrow C$"? Introducing errors is a distribution shift compared to what we want the model to generate during test time (i.e. a solution without error). Moreover, if we mask out the error part $[A \Rightarrow B$, oh I made a mistake, actually$]$, would it just reduce to training on perfectly correct data, with the model simply learning $A \Rightarrow C$?

In this paper, we provide initial results towards understanding the usefulness of including training data in the language model's corpus that teaches it to correct errors immediately. Specifically, we ask the following question:

> Can training on retry data (errors and immediate corrections) successfully teach the language model to perform error correction? **Can a language model achieve higher reasoning accuracy compared to training on the <u>same amount</u> of perfectly error-free data**?

To initiate our study and perform a controlled experiment, we need a setting where we can reliably collect data with errors and corrections. While it is tempting to use language models such as GPT-4 to synthesize such data, there is no evidence that GPT-4 can robustly generate errors or make those corrections consistently.[3] To avoid complicating our conclusions with the success rate and reliability of GPT-4, we need a different setting where we can 100% reliably generate errors and corrections.

In this work, we choose to use the iGSM dataset [29], which consists of a large number of program-generated (not LLM-generated) elementary-school level math reasoning problems. We discuss the dataset and provide examples in Section 2 to make this paper self-contained. Due to the generation procedure, we can also easily construct erroneous steps (simulating the reasoning mistakes that GPT-4 can make on such problems) and make 100% correct corrections to those steps. We choose this setting because mathematical reasoning errors are among the most widely observed errors made by language models. We outline our results in the following section.

---

[1] If they make an error when writing an article and realize it afterward, they will simply delete the error line in the final article.

[2] For example, we can have training data like [MaskStart] A wrong math solution [MaskEnd] What is the error in the above solution? ...

[3] Although GPT-4 can sometimes "verify" the correctness of a generated solution, its verification accuracy is far from 100%. In the example of Figure 1, using prompts such as "Please verify carefully and correct your solution," "Please check your solution line by line carefully, matching it to the problem description," or even "You made a mistake in Line 1, please correct it," or "I am very sure you made a mistake in Line 1, please correct it by checking it against the problem statement carefully," GPT-4o can insist that its solution is correct and makes no correction (using the 2024-08-06 version with temperature 0).

**Section 3: retry upon regret.** We begin with a warmup result. After pretraining the model on perfectly *error-free* math data, one can finetune it with a small number of trainable parameters (e.g., a rank-8 update on its embedding layer) to detect errors in its solution steps. This finetuning process is remarkably lightweight and essentially indicates that the model "already knows" it has made a mistake (i.e., its internal states exhibit regretful behavior).[4]

Next, we consider a "retry upon regret" generation process. During the inference phase, if the model detects an error in its generated solution step, it will "regenerate" from the end of its previous sentence. We demonstrate that this process leads to improved accuracy on this math dataset, surpassing what can be achieved using beam search.

*Remark* 1.1. This should not be confused with the "self-correction" works [11, 15, 19], where the model evaluates its own output through careful prompting. Here, we finetune the pretrained model for error detection ($>99\%$ accuracy), and this error detection accuracy is significantly higher than that achieved through prompting.

While this provides strong evidence that language models "regret" their mistakes, improving final accuracy relies solely on randomness (for re-generation) to correct errors — similar to beam search.[5] This can take many rounds to re-generate, which is inefficient and requires the error detector to be highly accurate. It also alters the generation process, which might not align with the "general intelligence" framework, where the goal is typically for one model to perform all tasks using the same (autoregressive) decoding algorithm.

**Section 4: pretrain with retry data.** We now turn our attention to *retry data*. If the pretraining data includes errors and their immediate corrections (which we call "retry data"), the model can learn to not only detect errors but also correct them, achieving much higher accuracy compared to "retry upon regret" in Section 3. This differs from the "self-correction" line of work: the model learns to *automatically retry* after detecting an error, without requiring additional prompting or multi-stage generation.

Furthermore, we discover that even when the model is pretrained entirely on retry data with high error rate (e.g., for the iGSM data, $p = 20\%$ or even $p = 50\%$, meaning roughly half of the solution steps have an inserted error), it does not tend to produce erroneous steps during generation. The model still strives to generate "perfectly correct" solutions most of the time and only corrects itself on the rare occasions when it makes a mistake, leading to an overall improvement in accuracy. In fact, within a reasonable range, the higher the $p$ value in the pretraining data, the better the model performs in reasoning accuracy.[6] We also observe that it is not necessary to perform label masking on the errors, so the vanilla autoregressive (causal masking) training simply works.

**Section 5: finetune with retry data.** In contrast, if a model is already pretrained with error-free data, even using sufficiently many (additional) retry data, with a wide range of LoRA finetuning configurations, label masking or not, the model's accuracy does not significantly improve. This indicates that the skill of error correction can be very different from the original error-free reasoning, and thus requires major weight changes, potentially beyond what parameter-efficient fine-tuning (PEFT) can handle. For comparison, we show that full finetuning is effective when sufficiently many retry data is available, though this resembles continued pretraining.

---

[4]Similar observations have been made in [13, 14, 26], where a pretrained transformer can be easily finetuned to effectively verify its own solutions (true or false).

[5]Beam search does not have the "error detection" subroutine but uses the model's next-token prediction probability distribution to simulate a re-generation process.

[6]To provide a strong comparison, for different $p$ we pretrain over the same number of tokens; so a higher $p$ means we pretrain over a smaller number of problems, because solutions with a larger $p$ are longer.

Thus, unlike error detection (see Section 3, where even a rank-8 adaptation on the embedding layer suffices), **error correction** is **not** a skill that can be easily adapted from a model pretrained with only error-free data. This implies that retry data should be included in the pretraining stage for practical LLM training, rather than in the finetuning (alignment) stage.

**Section 6: prepare fake retry data.** Retry data can be difficult to obtain, so we explore practical methods to automatically augment correct math solutions with "fake" mistakes, ideally without needing to semantically parse or understand the correct solution. The most effective method we found is to introduce a random future step $B$ as a "fake error" at each step $A$ in the solution, followed by $A$ as its "correction." This approach encourages the model not to skip steps, even though some solution steps may be interchangeable in order so these may not be true errors. In the synthetic iGSM setting, this method is nearly as effective as retry data with perfectly designed errors and corrections, and it could potentially be adapted to real-life math problems.

**Conclusion.** The full conclusion is deferred to Section 7. By utilizing fully-controllable synthetic data (e.g., controlling error rates or label masking), conducting controlled experiments (e.g., beam search vs. retry vs. error-free; pretrain vs. finetune), and performing fair comparisons (e.g., same number of training tokens), the goal of this paper is to try to predict the needs of future LLMs. We do not claim that the synthetic data used here can directly aid in building future LLMs. However, given that commercial LLMs already employ synthetic data [16, 25] and future LLMs are rumored to use $Q^\star$, it is perhaps crucial to understand how to best prepare and use such data effectively to teach models to learn from the mistakes.

## 2 Synthetic Math Data From Prior Work

Ye et al. [29] introduced a family of controllable, synthetic datasets of math problems with step-by-step solutions. These data simulate GSM8K [9], while removing arithmetic difficulties (by restricting computations to integers modulo 23) and common sense knowledge (e.g., a candle burns and its length decreases). What remains is the "logic reasoning" part. The dataset has much larger diversity (over 90 trillion solution templates), and the solutions are fully verifiable. We briefly summarize it to make the paper self-contained, emphasizing some important aspects.

An example from their dataset is in Figure 2. The structure graph describes the set of *instance parameters*, such as "the number of school daypacks in each film studio." They also allow for *abstract parameters*, such as "the (total) number of backpacks in central high," which requires hierarchical computation.[7]

**The exact data construction is not important for this paper.** What matters is that the parameters form a *dependency graph*, as shown in Figure 2, where a parameter can be computed only when its predecessors have all been computed. To remove arithmetic difficulty, the computations are broken into binary operations — such as $12 + 13 + 7$ is broken into $(12 + 13) + 7$— ensuring that failure to solve the problems is not due to arithmetic difficulties. They use op to denote the number of operations needed in the solution and prepared four families of data:

---
[7]In this example, it equals $A \times (B_1 + B_2)$ for $A$ = central high's number of film studios, $B_1, B_2$ = each film studio's number of school daypacks / messenger backpacks.

> **(Problem)** The number of each Riverview High's Film Studio equals 5 times as much as the sum of each Film Studio's Backpack and each Dance Studio's School Daypack. The number of each Film Studio's School Daypack equals 12 more than the sum of each Film Studio's Messenger Backpack and each Central High's Film Studio. The number of each Central High's Film Studio equals the sum of each Dance Studio's School Daypack and each Film Studio's Messenger Backpack. The number of each Riverview High's Dance Studio equals the sum of each Film Studio's Backpack, each Film Studio's Messenger Backpack, each Film Studio's School Daypack and each Central High's Backpack. The number of each Dance Studio's School Daypack equals 17. The number of each Film Studio's Messenger Backpack equals 13. *How many Backpack does Central High have?*

> **(Solution)** Define Dance Studio's School Daypack as p; so p = 17. Define Film Studio's Messenger Backpack as W; so W = 13. Define Central High's Film Studio as B; so B = p + W = 17 + 13 = 7. Define Film Studio's School Daypack as g; R = W + B = 13 + 7 = 20; so g = 12 + R = 12 + 20 = 9. Define Film Studio's Backpack as w; so w = g + W = 9 + 13 = 22. Define Central High's Backpack as c; so c = B * w = 7 * 22 = 16. *Answer: 16.*

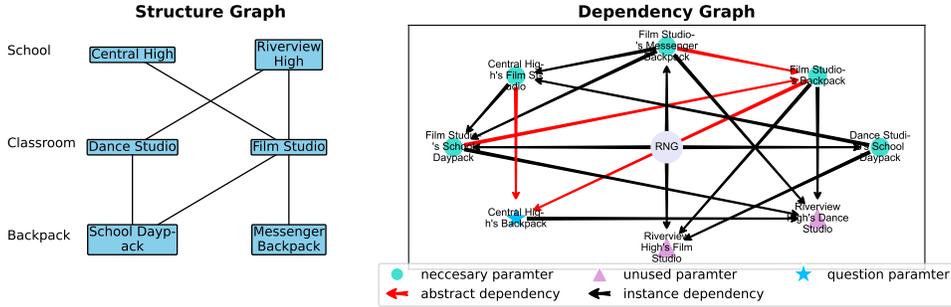

Figure 2: An example of a math problem with op = 7 operations needed to compute its solution. A much harder example with op = 20 is in Figure 9.

- iGSM-med$_{pq/qp}$ uses op $\leq 15$ for train; op $\in \{20, 21, 22, 23\}$ for OOD (out-of-distribution) test.
- iGSM-hard$_{pq/qp}$ uses op $\leq 21$ for training and op $\in \{28, 29, 30, 31, 32\}$ for OOD testing.

Here, $pq$ denotes the problem description comes before the question, and $qp$ otherwise. They also introduce reask data for evaluation purposes: for instance, iGSM-med$_{pq}^{\text{op}=20,\text{reask}}$ is a dataset constructed by first creating iGSM-med$_{pq}^{\text{op}=20}$ and then re-sampling a parameter to query — this greatly changes their math data distribution, making it a good candidate for OOD evaluation. (After reask, the problem's op value may change.)

They showed that GPT-4/GPT-4o cannot solve such problems for op $> 10$ (even with few-shot learning and their best efforts to remove English and arithmetic difficulties), indicating that the datasets are of some non-trivial difficulty. For completeness, we include in Figure 9 an example with op = 21 to illustrate that these problems require non-trivial reasoning even for humans.

## 3 Result 0-1: Language Models Can Retry Upon Regret

Generative models solve math problems step by step in a chain-of-thought (CoT) manner. Each step in our math problems is a single sentence formatted as "Define [param] as X; so...", as shown in Figure 2. How do generative models make mistakes in this CoT process?

The most *common* reasoning mistake[8] occurs when the model generates a [param] that is not yet ready for computation (i.e., the model has not determined the values of all the parameters that [param] depends on, also known as "skipping steps", a frequent error even in GPT-4 [8]).

---

[8]In practice, language models can make other mistakes such as in arithmetic or common sense (see Section 2); however, the design of the iGSM datasets has removed such difficulties, allowing us to focus solely on the reasoning aspect.

For example, in Figure 2, if the model generates "Define Central High's Film Studio" as the first few words in its solution, it cannot go back and erase this sentence, leading to a mistake. Ye et al. [29] confirmed that not only do language models pretrained using the iGSM datasets make mistakes in this manner, but even GPT-4/GPT-4o make such mistakes (when using few-shot). It's worth noting that the failure example in Figure 1 is also in this spirit.

### 3.1 Result 0: Models Can Be "Regretful" After Making Mistakes

Interestingly, their same paper also implies the following:

> **Result 0** (corollary of [29]). *For models pretrained on iGSM (with correct solutions only!), during their solution generation process, after writing "Define [param] as" for a wrong [param], they often "realize" such a mistake, showing a regretful pattern in their internal states.*

To see this, one can apply their probing technique (illustrated in Figure 3(a)) to extract information from the model's last hidden layer after "Define [param $A$] as" to see if the model knows $A$ can truly be computed next. This probing task is denoted as $\mathtt{can\_next}(A) \in \{\text{true}, \text{false}\}$. They found:

- When $A$ ranges over all possible parameters, the probing 99% accurately predicts $\mathtt{can\_next}(A)$, meaning the model knows if $A$ can be computed next, even for the hardest $\mathtt{op} = 32$ problems.

- When the model makes a mistake, the first sentence with a mistake usually has $\mathtt{can\_next}(A) = \text{false}$. Probing shows the model has $\sim 60\%$ chance of knowing $\mathtt{can\_next}(A) = \text{false}$, indicating it often knows it has made a mistake, *right after* stating the parameter name in full.[9]

The statistical difference between the two cases signifies that the model's internal states do exhibit a "regretful" pattern, which can be detected via probing such as $\mathtt{can\_next}$, which is almost just a linear classifier on top of its hidden states.[10] In other words, **error detection is easy** and is a skill almost already embedded within the model's internal states, even when pretrained on correct math problems only.

### 3.2 Result 1: Let Models Retry Upon Regret

If a model knows it is a mistake, why does it generate the wrong [param $A$] in the first place? The issue lies in the generation process. *Before explicitly stating* "Define [param $A$] as", the model might falsely think $A$ is ready to compute among *all the parameters* it can focus on. After stating it, the model shifts its focus to the actual computation of $A$, and this is the moment it can better realize that $A$ is not ready for computation (using its attention mechanism).[11]

Now that we know the model exhibits some "regret" towards the mistake, can we use this to improve accuracy?

**Retry upon regret.** We conducted an experiment using the probing result to guide the model's generation process. After generating each solution sentence, we use the $\mathtt{can\_next}$ probing to determine if the model knows it has made a mistake. If so, we revert to the end of the previous sentence and regenerate. We use multinomial sampling (i.e., beam=1 and dosample=true) during this regeneration process, with a maximum of 10 total retries for generating each solution.[12] We

---

[9]Note that 60% accuracy is significant. If it were a random guess, 50% accuracy would be trivial. However, the probing method is 99% accurate in predicting true or false, and only on a small set of examples (i.e., when making mistakes), it has a 60% chance of correctly predicting $\mathtt{can\_next}(A) = \text{false}$.

[10]This aligns with observations that detecting mistakes is usually easy: in works like [13, 14, 26], they show pretrained transformers can be easily fine-tuned to effectively verify their own solutions (true or false).

[11]Similar phenomenon also occurs in knowledge partial retrieval [3].

[12]It is necessary to limit the maximum number of retries to avoid infinite loops.

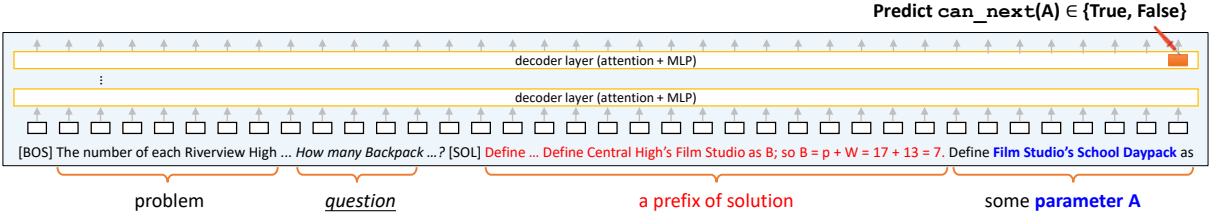

(a) `can_next` probing [29]. After pretraining, V-probing technique can detect if the model's internal states exhibit a regretful pattern: right after "Define param $A$ as," the model no longer thinks $A$ is ready for compute.

(b) Retry upon regret vs. original accuracies. Version1/2 uses the `can_next` probing to detect regret (see Section 3.2, they can detect errors to 99% accuracy), and versionP uses a perfect error detector to decide when to retry.

Figure 3: The `can_next` probing and using it to assist the model's generation process.

call this method "retry upon regret", and we carefully compare a few different versions (details are in Appendix D.2).

- The original `can_next` probing [29] freezes the pretrained model and fine-tunes a rank-8 update on the embedding layer — we call this "version1", and it can 99% accurately detect errors.[13]

- We additionally consider a different `can_next` probing that fine-tunes more parameters but with only 0.25% data — we call this "version2", and it can also 99% accurately detect errors.

- For comparison, we also consider the perfect, ground-truth `can_next` error detector and call this "versionP"; and we use "versionP50" to indicate using a maximum of 50 total retries.

The **goal** of *version1* is to check if we can use extremely small parameter changes from the pretrained model for error detection and use this towards higher accuracies. The **goal** of *version2* is to check if we can use a small number of fine-tuning samples for error detection. The **goal** of *versionP/P50* is for comparison purposes; after all, allowing perfect error detection can enable the model to correct nearly all mistakes if retried infinitely.

We compared this with the original generation (with beam=1, 4, 16, and 32); and we used beam=1 or 4 for all versions of "retry upon regret." The results give rise to the following:

> **Result 1** (Figure 3(b)). *The idea of "retry upon regret" can improve reasoning accuracy, to an extent better than beam search. However, the accuracy gain can be limited unless the error detector is highly accurate. This also alters the inference process and increases the inference complexity.*

We explain the three sentences in detail below:

---
[13]See their reported probing accuracy table, where the `can_next`($A$) probing accuracies can be 99% even for op = 23 on iGSM-med or for op = 32 on iGSM-hard.

> **(Solution - retry rate 0.5)** Define Dance Studio's School Daypack as p; so p = 17. ~~Define Film Studio's School Daypack as~~ [BACK]. Define Film Studio's Messenger Backpack as W; so W = 13. ~~Define Central High's Classroom as~~ [BACK]. ~~Define Central High's Backpack as~~ [BACK]. Define Central High's Film Studio as B; so B = p + W = 17 + 13 = 7. Define Film Studio's School Daypack as g; R = W + B = 13 + 7 = 20; so g = 12 + R = 12 + 20 = 9. ~~Define Riverview High's Dance Studio as~~ [BACK]. Define Film Studio's Backpack as w; so w = g + W = 9 + 13 = 22. ~~Define Riverview High's Dance Studio as~~ [BACK]. Define Central High's Backpack as c; so c = B * w = 7 * 22 = 16.

(a) A solution example identical to Figure 2 but with retry_rate = 0.5. The strikethrough like "~~Define Central High's Backpack as~~" is for illustration purpose, and the actual data is normal English text without strikethrough symbols.

| | iGSM-med_pq | | iGSM-med_qp | | iGSM-hard_pq | | iGSM-hard_qp | | row |
|---|---|---|---|---|---|---|---|---|---|
| | in-dist | out-of-dist (OOD) | in-dist | out-of-dist (OOD) | in-dist | out-of-dist (OOD) | in-dist | out-of-dist (OOD) | avg |
| original | 99.9 99.1 | 92.0 88.4 84.5 77.7 91.6 | 100 99.3 | 92.4 89.9 84.8 78.3 91.4 | 100 99.4 | 94.9 93.7 91.4 88.4 84.0 91.4 | 100 99.4 | 94.5 93.3 91.8 88.3 85.3 89.4 | 91.8 |
| retry \| retryrate0.05 | 100 99.7 | 94.8 93.0 88.5 83.9 94.5 | 100 99.4 | 93.3 89.2 85.7 78.8 92.4 | 100 99.6 | 96.6 94.8 93.2 90.7 86.9 94.8 | 100 99.7 | 96.2 93.7 92.7 89.4 86.3 92.1 | 93.3 |
| retry \| retryrate0.05 (with mask) | 100 99.5 | 93.1 89.5 84.6 78.5 93.2 | 100 99.6 | 95.2 91.3 88.4 82.3 92.6 | 100 99.8 | 97.3 95.2 94.2 91.6 88.6 94.6 | 100 99.6 | 95.7 94.1 91.7 88.5 84.9 90.0 | 93.1 |
| retry \| retryrate0.1 | 100 99.7 | 97.1 95.8 93.3 90.6 95.2 | 99.9 99.6 | 94.0 90.6 87.6 81.7 91.9 | 100 99.7 | 97.1 96.2 93.8 91.2 87.2 96.3 | 100 99.9 | 98.1 97.4 95.8 93.2 90.4 93.4 | 94.9 |
| retry \| retryrate0.1 (with mask) | 100 99.8 | 96.6 94.2 91.5 87.5 96.9 | 100 99.8 | 96.0 93.8 89.9 84.0 92.5 | 100 99.9 | 97.3 96.6 94.6 91.8 87.9 92.9 | 100 99.9 | 97.2 96.0 94.8 91.9 89.2 92.4 | 94.8 |
| retry \| retryrate0.2 | 100 99.9 | 97.9 96.8 95.2 91.6 96.0 | 100 99.8 | 96.7 94.5 92.0 88.3 93.4 | 100 99.8 | 97.9 96.6 95.8 93.7 91.7 95.6 | 100 99.9 | 98.4 97.0 96.4 95.0 92.1 92.9 | 96.2 |
| retry \| retryrate0.2 (with mask) | 100 99.9 | 97.2 95.5 93.0 87.8 95.8 | 100 99.9 | 97.4 96.2 94.4 90.9 93.7 | 100 99.9 | 98.6 97.1 96.5 95.0 92.9 95.3 | 100 99.9 | 98.5 97.5 96.8 95.3 94.0 91.3 | 96.3 |
| retry \| retryrate0.4 | 100 99.8 | 97.6 95.8 94.3 90.6 97.6 | 100 99.9 | 97.7 95.4 93.4 90.2 93.4 | 100 99.9 | 99.1 98.1 97.8 96.7 94.8 97.3 | 100 99.8 | 97.1 95.4 94.3 93.0 89.3 93.1 | 96.4 |
| retry \| retryrate0.5 | 100 99.9 | 98.3 96.8 95.8 93.6 96.1 | 100 99.9 | 98.7 98.0 96.9 94.5 96.0 | 100 99.9 | 98.9 98.1 97.1 95.6 93.9 98.8 | 100 99.9 | 98.3 97.2 96.3 94.9 93.7 97.0 | 97.5 |
| retry \| retryrate0.5 (with mask) | 100 100 | 98.8 97.9 96.8 94.8 96.0 | 100 99.8 | 97.9 96.9 95.2 93.2 91.5 | 100 100 | 99.1 98.9 98.4 96.6 96.2 97.3 | 100 100 | 99.5 98.9 98.4 97.5 95.8 96.9 | 97.7 |

Column x-axis labels: op ≤ 15, op = 15, op = 20, op = 21, op = 22, op = 23, op = 20 (reask) | op ≤ 15, op = 15, op = 20, op = 21, op = 22, op = 23, op = 20 (reask) | op ≤ 21, op = 21, op = 28, op = 29, op = 30, op = 31, op = 32, op = 28 (reask) | op ≤ 21, op = 21, op = 28, op = 29, op = 30, op = 31, op = 32, op = 28 (reask)

(b) Comparison of models pretrained using iGSM data with retry_rate > 0. For a stronger comparison, the model is pretrained on the retry vs original (error-free) data using the same number of tokens (i.e., retry data has fewer problems than original data) and identical training parameters, see Appendix D.1.

Figure 4: Pretrain language models on error-free vs retry data. **Observation:** especially on the hardest tasks (op = 23 or 32), models pretrained from retry data exhibit the greatest improvements for larger retry_rate.

- Comparing version1/2 with beam32, we see "retry upon regret" improves upon beam search. However, even though error detection is 99% accurate, this improvement is still marginal: about 2% for the op = 23 (resp. op = 32) case for iGSM-med (resp. iGSM-hard).

- Comparing version1/2 with versionP, we see that the success of "retry upon regret" largely depends on an extremely accurate error detector — increasing the error detection success rate from 99% to 100% can significantly improve the final reasoning accuracy, but this is too ideal.[14]

- The idea of "retry upon regret" increases the inference complexity because one needs to keep an error detector model alongside and keep checking the correctness of the generated solution steps. In the event of an error, the model needs to regenerate using randomness (possibly multiple times) until it passes the error detector. Ideally, one wishes to have just a single model to achieve "general intelligence" using the simplest autoregressive decoding algorithm, without multi-round error corrections.

## 4 Result 2-6: Pretrain with Retry Data

In this section, we prepare pretrain data to teach the model to directly correct mistakes.

**Math data with retry.** Since we use a controllable, synthetic math dataset, we can, at the beginning of each solution sentence, with probability retry_rate ∈ [0, 1), insert a wrong parameter that cannot be computed next, followed by a special token [BACK].[15] We repeat this process, so with probability (retry_rate)$^2$, it may generate another wrong parameter at the same location, and so on. We provide an extreme example with retry_rate = 0.5 in Figure 4(a), and a more complex

---

[14]After all, a false negative in error detection results in a wrong answer, and having a false positive can result in the model regenerating too many times.

[15]This parameter is uniformly randomly chosen from all such parameters, except those already appearing.

## Figure 5

|  | iGSM-med_pq | | | | | | | iGSM-med_qp | | | | | | | iGSM-hard_pq | | | | | | | iGSM-hard_qp | | | | | | |
| --- | --- | --- | --- | --- | --- | --- | --- | --- | --- | --- | --- | --- | --- | --- | --- | --- | --- | --- | --- | --- | --- | --- | --- | --- | --- | --- | --- | --- |
|  | in-dist | | out-of-dist (OOD) | | | | | in-dist | | out-of-dist (OOD) | | | | | in-dist | | out-of-dist (OOD) | | | | | in-dist | | out-of-dist (OOD) | | | | |
|  | op≤15 | op=15 | op=20 | op=21 | op=22 | op=23 | op=20 (reask) | op≤15 | op=15 | op=20 | op=21 | op=22 | op=23 | op=20 (reask) | op≤21 | op=21 | op=28 | op=29 | op=30 | op=31 | op=32 | op=28 (reask) | op≤21 | op=21 | op=28 | op=29 | op=30 | op=31 | op=32 | op=28 (reask) |
| retry \| retryrate0.05 | 0.00 | 0.01 | 0.03 | 0.04 | 0.05 | 0.06 | 0.18 | 0.00 | 0.01 | 0.03 | 0.05 | 0.05 | 0.07 | 0.06 | 0.00 | 0.00 | 0.04 | 0.04 | 0.05 | 0.07 | 0.08 | 0.25 | 0.01 | 0.01 | 0.04 | 0.05 | 0.06 | 0.08 | 0.10 | 0.13 |
| retry \| retryrate0.05 (with mask) | 0.00 | 0.01 | 0.03 | 0.03 | 0.06 | 0.08 | 0.10 | 0.00 | 0.01 | 0.03 | 0.04 | 0.06 | 0.07 | 0.05 | 0.00 | 0.01 | 0.05 | 0.07 | 0.08 | 0.10 | 0.13 | 0.17 | 0.00 | 0.01 | 0.04 | 0.07 | 0.06 | 0.08 | 0.10 | 0.07 |
| retry \| retryrate0.1 | 0.00 | 0.01 | 0.05 | 0.07 | 0.09 | 0.15 | 0.17 | 0.00 | 0.00 | 0.02 | 0.06 | 0.08 | 0.08 | 0.12 | 0.00 | 0.01 | 0.06 | 0.09 | 0.11 | 0.12 | 0.14 | 0.50 | 0.00 | 0.01 | 0.06 | 0.06 | 0.08 | 0.08 | 0.09 | 0.17 |
| retry \| retryrate0.1 (with mask) | 0.00 | 0.01 | 0.10 | 0.15 | 0.19 | 0.23 | 0.12 | 0.00 | 0.01 | 0.05 | 0.04 | 0.05 | 0.08 | 0.06 | 0.01 | 0.02 | 0.05 | 0.10 | 0.11 | 0.10 | 0.15 | 0.35 | 0.00 | 0.01 | 0.04 | 0.07 | 0.08 | 0.11 | 0.12 | 0.08 |
| retry \| retryrate0.2 | 0.00 | 0.02 | 0.10 | 0.11 | 0.17 | 0.22 | 0.19 | 0.00 | 0.01 | 0.08 | 0.11 | 0.17 | 0.20 | 0.11 | 0.01 | 0.02 | 0.10 | 0.15 | 0.16 | 0.19 | 0.25 | 0.38 | 0.00 | 0.02 | 0.10 | 0.13 | 0.17 | 0.22 | 0.26 | 0.19 |
| retry \| retryrate0.2 (with mask) | 0.00 | 0.01 | 0.11 | 0.16 | 0.23 | 0.29 | 0.25 | 0.00 | 0.01 | 0.11 | 0.13 | 0.20 | 0.22 | 0.12 | 0.00 | 0.01 | 0.05 | 0.07 | 0.09 | 0.13 | 0.19 | 0.31 | 0.00 | 0.01 | 0.04 | 0.09 | 0.09 | 0.12 | 0.13 | 0.27 |
| retry \| retryrate0.4 | 0.02 | 0.10 | 0.27 | 0.36 | 0.45 | 0.59 | 0.37 | 0.02 | 0.12 | 0.38 | 0.49 | 0.65 | 0.82 | 0.67 | 0.03 | 0.24 | 0.51 | 0.57 | 0.62 | 0.74 | 0.83 | 0.85 | 0.03 | 0.23 | 0.54 | 0.68 | 0.73 | 0.89 | 1.02 | 0.79 |
| retry \| retryrate0.5 | 0.26 | 0.93 | 1.85 | 2.15 | 2.41 | 2.86 | 0.78 | 0.22 | 0.73 | 1.50 | 1.62 | 1.92 | 2.32 | 0.58 | 0.31 | 1.36 | 2.37 | 2.54 | 2.71 | 2.98 | 3.19 | 1.20 | 0.35 | 1.43 | 2.53 | 2.72 | 2.93 | 3.38 | 3.70 | 1.26 |
| retry \| retryrate0.5 (with mask) | 0.00 | 0.01 | 0.14 | 0.18 | 0.29 | 0.34 | 0.42 | 0.00 | 0.01 | 0.09 | 0.16 | 0.23 | 0.31 | 0.14 | 0.00 | 0.02 | 0.12 | 0.14 | 0.20 | 0.21 | 0.24 | 0.81 | 0.00 | 0.01 | 0.08 | 0.11 | 0.10 | 0.19 | 0.23 | 0.20 |

retry counts (on correct)

(a) Retry counts on correct solutions

|  | iGSM-med_pq | | | | | | iGSM-med_qp | | | | | | iGSM-hard_pq | | | | | | iGSM-hard_qp | | | | | |
| --- | --- | --- | --- | --- | --- | --- | --- | --- | --- | --- | --- | --- | --- | --- | --- | --- | --- | --- | --- | --- | --- | --- | --- | --- |
|  | in-dist | out-of-dist (OOD) | | | | | in-dist | out-of-dist (OOD) | | | | | in-dist | out-of-dist (OOD) | | | | | in-dist | out-of-dist (OOD) | | | | |
|  | op≤15 | op=15 | op=20 | op=21 | op=22 | op=23 | op=20 (reask) | op≤15 | op=15 | op=20 | op=21 | op=22 | op=23 | op=20 (reask) | op≤21 | op=21 | op=28 | op=29 | op=30 | op=31 | op=32 | op=28 (reask) | op≤21 | op=21 | op=28 | op=29 | op=30 | op=31 | op=32 | op=28 (reask) |
| retry \| retryrate0.05 | | 0.68 | 0.63 | 0.70 | 0.64 | 1.37 | | 0.82 | 1.20 | 1.05 | 0.94 | 0.59 | | 1.57 | 1.26 | 1.23 | 0.92 | 1.29 | 2.88 | | 0.93 | 1.04 | 0.89 | 1.35 | 0.97 | 1.76 |
| retry \| retryrate0.05 (with mask) | | 1.00 | 1.17 | 1.13 | 1.03 | 1.41 | | 0.90 | 1.00 | 1.15 | 1.04 | 0.41 | | 1.20 | 1.44 | 0.86 | 1.01 | 1.02 | 2.30 | | 1.45 | 1.21 | 1.04 | 1.17 | 0.97 | 0.69 |
| retry \| retryrate0.1 | | 1.33 | 1.19 | 1.19 | 1.20 | 2.26 | | 0.96 | 1.08 | 0.62 | 0.75 | 0.76 | | 1.64 | 1.57 | 1.73 | 1.48 | 1.34 | 3.88 | | 0.11 | 0.11 | 0.10 | 0.20 | 0.25 | 0.53 |
| retry \| retryrate0.1 (with mask) | | 3.79 | 3.24 | 2.85 | 2.67 | 5.05 | | 1.05 | 0.90 | 0.84 | 0.88 | 0.93 | | 1.68 | 2.36 | 2.30 | 1.91 | 1.56 | 4.99 | | 1.53 | 2.02 | 1.48 | 1.61 | 1.32 | 2.01 |
| retry \| retryrate0.2 | | 1.62 | 2.56 | 2.39 | 1.82 | 2.30 | | 1.56 | 1.29 | 1.54 | 1.52 | 1.86 | | 1.10 | 1.73 | 1.18 | 1.28 | 1.18 | 2.22 | | 0.89 | 0.64 | 1.81 | 0.99 | 1.47 | 1.35 |
| retry \| retryrate0.2 (with mask) | | 1.83 | 2.05 | 2.51 | 2.17 | 4.20 | | 1.58 | 1.72 | 3.23 | 2.19 | 1.75 | 1.19 | | 2.61 | 1.88 | 3.23 | 2.28 | 2.31 | 2.46 | | 0.20 | 0.18 | 0.24 | 0.36 | 0.36 | 3.06 |
| retry \| retryrate0.4 | | 0.62 | 0.72 | 1.07 | 1.14 | 1.63 | | 0.81 | 1.14 | 1.05 | 1.54 | 2.21 | | 3.29 | 2.96 | 2.84 | 3.47 | 2.89 | 3.68 | | 0.67 | 0.75 | 0.89 | 1.09 | 1.21 | 1.97 |
| retry \| retryrate0.5 | | 4.82 | 6.25 | 5.27 | 5.90 | 8.51 | | 2.11 | 2.46 | 2.63 | 2.99 | 1.40 | | 3.67 | 3.26 | 3.80 | 3.25 | 3.97 | 3.76 | | 5.15 | 7.50 | 5.76 | 5.47 | 8.05 | 6.01 |
| retry \| retryrate0.5 (with mask) | | 3.81 | 4.70 | 4.75 | 4.69 | 5.00 | | 0.28 | 0.49 | 0.51 | 0.74 | 1.80 | | 3.02 | 0.98 | 1.84 | 1.42 | 2.42 | 9.32 | | 1.38 | 0.58 | 1.85 | 2.08 | 2.75 | 0.94 |

retry counts (on wrong)

(b) Retry counts on wrong solutions (in-dist cases are ignored since there are too few wrong solutions)

Figure 5: Model's average number of retries per test problem. Details see Appendix D.1.

---

example with op = 21 and retry_rate = 0.2 in Figure 9. We call this the **retry data** (with error and correction), to distinguish that from the original (error-free) data.

- *Can a language model be pretrained using retry data to improve its reasoning accuracy compared with language models pretrained on the error-free data?*

We conduct experiments on the iGSM-med$_{pq/qp}$ and iGSM-hard$_{pq/qp}$ datasets for retry_rate ranging from 0.01 to 0.5. **For a controlled comparison**, we compare models when they are pretrained using the *same number of tokens* (along with other parameters).[16] This means, when pretrained with retry data, the model sees *fewer* math problems compared to the original training; when pretrained with larger retry_rate, the model also sees fewer math problems.

Another important question when pretraining with retry data is that:

- *Is it necessary to introduce label masks to prevent the model from learning the mistakes?*

Adding masking might make intuitive sense because we do not want the model to learn the mistakes — we only want it to learn to correct errors if needed. To answer this, we also perform controlled experiment to compare (1) standard auto-regressive pretraining and (2) pretraining with masking: adding label masks to ignore the auto-regressive loss on the wrong parameter tokens (i.e., ignoring text with ~~strikethrough~~ in Figure 4(a)).

> **Result 2-3.** *Our results in Figure 4(b) strongly support that:*
>
> - *Within a reasonable range,[17] **the more mistakes the better.** Especially on hard problems, such as on iGSM-med$_{qp}^{op=23}$, the accuracy jumps from 78% to 94% by using* retry_rate = 0.5.
>
> - ***Masking mistakes is unnecessary.*** *We observe that it is generally not needed to introduce label masking on the error data even for large* retry_rate = 0.5.

A reader seeing this for the first time may find the above results unbelievable: if the pretrain

---

[16]The same training parameters such as batch size, total steps, learning rates, see Appendix D. In particular, we adopt the learning rate from [29] (which was best tuned for the original error-free data) directly to the retry data.

[17]Naturally, retry_rate cannot approach 1. Exploring such extreme failure settings is not particularly interesting. For instance, retry_rate = 0.5 is already sufficiently extreme, indicating that half of the solution steps contain errors.

| | iGSM-med_pq | | iGSM-med_qp | | iGSM-hard_pq | | iGSM-hard_qp | |
|---|---|---|---|---|---|---|---|---|
| | in-dist | out-of-dist (OOD) | in-dist | out-of-dist (OOD) | in-dist | out-of-dist (OOD) | in-dist | out-of-dist (OOD) |
| unnecessary op \| original | 0.00 0.00 | 0.00 0.00 0.00 0.00 0.15 | 0.00 0.00 | 0.00 0.00 0.00 0.00 0.17 | 0.01 0.00 | 0.00 0.00 0.00 0.00 0.52 | 0.00 0.00 | 0.00 0.00 0.00 0.00 0.45 |
| unnecessary op \| retry \| retryrate0.05 | 0.01 0.00 | 0.00 0.00 0.00 0.00 0.18 | 0.00 0.00 | 0.00 0.00 0.00 0.00 0.18 | 0.01 0.00 | 0.00 0.00 0.00 0.00 0.49 | 0.02 0.00 | 0.00 0.00 0.01 0.01 0.57 |
| unnecessary op \| retry \| retryrate0.05 (with mask) | 0.01 0.00 | 0.00 0.00 0.00 0.00 0.27 | 0.01 0.00 | 0.00 0.00 0.00 0.00 0.18 | 0.00 0.00 | 0.00 0.00 0.01 0.00 0.01 0.54 | 0.02 0.00 | 0.00 0.00 0.00 0.00 0.44 |
| unnecessary op \| retry \| retryrate0.1 | 0.00 0.00 | 0.00 0.00 0.00 0.00 0.20 | 0.01 0.00 | 0.00 0.00 0.00 0.00 0.20 | 0.01 0.00 | 0.00 0.00 0.00 0.00 0.58 | 0.01 0.00 | 0.00 0.00 0.00 0.00 0.49 |
| unnecessary op \| retry \| retryrate0.1 (with mask) | 0.00 0.00 | 0.01 0.01 0.01 0.01 0.19 | 0.00 0.00 | 0.00 0.00 0.00 0.00 0.16 | 0.01 0.00 | 0.00 0.00 0.01 0.01 0.46 | 0.01 0.00 | 0.00 0.00 0.00 0.00 0.44 |
| unnecessary op \| retry \| retryrate0.2 | 0.00 0.00 | 0.00 0.00 0.00 0.00 0.12 | 0.01 0.00 | 0.00 0.00 0.00 0.00 0.19 | 0.01 0.00 | 0.00 0.00 0.00 0.00 0.71 | 0.01 0.00 | 0.00 0.00 0.00 0.00 0.59 |
| unnecessary op \| retry \| retryrate0.2 (with mask) | 0.00 0.00 | 0.01 0.01 0.01 0.01 0.15 | 0.00 0.00 | 0.00 0.00 0.00 0.00 0.16 | 0.01 0.00 | 0.00 0.00 0.00 0.00 0.50 | 0.01 0.00 | 0.00 0.00 0.00 0.00 0.41 |
| unnecessary op \| retry \| retryrate0.4 | 0.01 0.00 | 0.00 0.00 0.00 0.00 0.17 | 0.01 0.00 | 0.00 0.00 0.00 0.00 0.17 | 0.01 0.00 | 0.00 0.00 0.00 0.00 0.54 | 0.01 0.00 | 0.00 0.00 0.00 0.00 0.51 |
| unnecessary op \| retry \| retryrate0.5 | 0.00 0.00 | 0.01 0.00 0.01 0.00 0.15 | 0.01 0.00 | 0.00 0.00 0.01 0.00 0.21 | 0.01 0.00 | 0.00 0.00 0.00 0.00 0.58 | 0.01 0.00 | 0.00 0.00 0.00 0.00 0.50 |
| unnecessary op \| retry \| retryrate0.5 (with mask) | 0.00 0.00 | 0.01 0.01 0.01 0.01 0.17 | 0.00 0.00 | 0.00 0.00 0.01 0.01 0.22 | 0.02 0.00 | 0.00 0.01 0.00 0.00 0.59 | 0.01 0.00 | 0.00 0.00 0.00 0.00 0.57 |
| unnecessary param \| original | 0.00 0.00 | 0.00 0.00 0.00 0.00 0.12 | 0.00 0.00 | 0.00 0.00 0.00 0.00 0.15 | 0.01 0.00 | 0.00 0.00 0.00 0.00 0.36 | 0.00 0.00 | 0.00 0.00 0.00 0.00 0.32 |
| unnecessary param \| retry \| retryrate0.05 | 0.01 0.00 | 0.00 0.00 0.00 0.00 0.13 | 0.00 0.00 | 0.00 0.00 0.00 0.00 0.14 | 0.01 0.00 | 0.00 0.00 0.00 0.00 0.33 | 0.02 0.00 | 0.00 0.00 0.00 0.00 0.40 |
| unnecessary param \| retry \| retryrate0.05 (with mask) | 0.01 0.00 | 0.00 0.00 0.00 0.00 0.21 | 0.01 0.00 | 0.00 0.00 0.00 0.00 0.14 | 0.00 0.00 | 0.00 0.00 0.01 0.00 0.00 0.38 | 0.01 0.00 | 0.00 0.00 0.00 0.00 0.32 |
| unnecessary param \| retry \| retryrate0.1 | 0.00 0.00 | 0.00 0.00 0.00 0.00 0.14 | 0.01 0.00 | 0.00 0.00 0.00 0.00 0.16 | 0.01 0.00 | 0.00 0.00 0.00 0.00 0.40 | 0.01 0.00 | 0.00 0.00 0.00 0.00 0.36 |
| unnecessary param \| retry \| retryrate0.1 (with mask) | 0.00 0.00 | 0.00 0.01 0.01 0.01 0.14 | 0.00 0.00 | 0.00 0.00 0.00 0.00 0.15 | 0.01 0.00 | 0.00 0.00 0.00 0.01 0.33 | 0.01 0.00 | 0.00 0.00 0.00 0.00 0.32 |
| unnecessary param \| retry \| retryrate0.2 | 0.00 0.00 | 0.00 0.00 0.00 0.00 0.09 | 0.01 0.00 | 0.00 0.00 0.00 0.00 0.15 | 0.01 0.00 | 0.00 0.00 0.00 0.00 0.48 | 0.01 0.00 | 0.00 0.00 0.00 0.00 0.43 |
| unnecessary param \| retry \| retryrate0.2 (with mask) | 0.00 0.00 | 0.00 0.01 0.00 0.01 0.11 | 0.01 0.00 | 0.00 0.00 0.00 0.00 0.19 | 0.01 0.00 | 0.00 0.00 0.00 0.00 0.35 | 0.01 0.00 | 0.00 0.00 0.00 0.00 0.30 |
| unnecessary param \| retry \| retryrate0.4 | 0.01 0.00 | 0.00 0.00 0.00 0.00 0.13 | 0.01 0.00 | 0.00 0.00 0.00 0.00 0.19 | 0.01 0.00 | 0.00 0.00 0.00 0.00 0.37 | 0.01 0.00 | 0.00 0.00 0.00 0.00 0.38 |
| unnecessary param \| retry \| retryrate0.5 | 0.00 0.00 | 0.00 0.00 0.00 0.00 0.12 | 0.01 0.00 | 0.00 0.00 0.00 0.00 0.17 | 0.01 0.00 | 0.00 0.00 0.00 0.00 0.40 | 0.01 0.00 | 0.00 0.00 0.00 0.00 0.36 |
| unnecessary param \| retry \| retryrate0.5 (with mask) | 0.00 0.00 | 0.01 0.01 0.01 0.01 0.13 | 0.00 0.00 | 0.00 0.00 0.01 0.01 0.18 | 0.02 0.00 | 0.00 0.00 0.00 0.00 0.41 | 0.01 0.00 | 0.00 0.00 0.00 0.00 0.42 |

Column headers (x-axis): op≤15, op=15, op=20, op=21, op=22, op=23, op=20 (reask) | op≤15, op=15, op=20, op=21, op=22, op=23, op=20 (reask) | op≤21, op=21, op=28, op=29, op=30, op=31, op=32, op=28 (reask) | op≤21, op=21, op=28, op=29, op=30, op=31, op=32, op=28 (reask)

(on correct solutions)

Figure 6: Model's average number of unnecessary operations or parameters (per test problem) used in its generated solutions (among correct solutions). Details see Appendix D.1.

data is full of mistakes (even with retry_rate = 0.5), doesn't this interfere with the model's learning and encourage it to make lots of mistakes? To address this, we compute the statistics on how many times the pretrained model performs retries (i.e., how many [BACK] tokens it uses) during evaluation. We discover that:

> **Result 4** (Figure 5). *Models pretrained on retry data hardly retry (unless retry_rate is very high).*

For instance, Figure 5(a) shows if retry_rate = 0.2, even when pretrained without label masking, the model retries an average of $< 0.3$ times even for math problems with large op. This is because at each solution step, the retry data still has a higher $1 - 0.2 = 0.8$ chance to produce an error-free step. Thus, a model pretrained from such data is still incentivized to generate correct steps (e.g., using a low temperature).[18] If retry_rate = 0.5, this average retry count becomes $2 \sim 4$, but can be mitigated using label masking. Only on those wrong solutions does the model try its best to retry and fail in the end, see Figure 5(b).

Another concern with the retry data is whether the model can still output shortest solutions. Ye et al. [29] discovered that language models can learn to produce shortest solutions without computing any unnecessary parameters.[19] Does the introduction of math data with errors increase the solution length? We discover that:

> **Result 5** (Figure 6). *Models pretrained on retry data can still output shortest math solutions.*

**Combining Results 2-5**, we see that it is safe to include retry data of this format as pretrain data to improve the model's reasoning accuracy. There is no change to the pretrain or inference process, no need to increase training time, and the solutions still satisfy many desirable properties.

Finally, it is tempting to compare retry data with beam search, which lets the model generate multiple solutions at each step and pick the most likely continuation based on the log-likelihood. As long as the model is pretrained on error-free data, even with 16 or 32 beams, its accuracy does not noticeable improve in all the cases (recall Figure 3(b)); while in contrast, if the model is pretrained

---

[18] Even when retry_rate = 0.5, if errors are sufficiently random (say, in a step there are 2 correct possibilities but 8 possible errors), the model is still incentivized to say correct sentences. This could be reminiscent of language model's learning on context-free grammars with mistakes [2]: even though the model is trained *only* using data with grammar mistakes, it can output sentences correctly respecting the grammar at lower temperature.

[19] They discover the model achieves so via non-trivial mental planning to precompute the set of necessary parameters, before it starts to even generate the first solution sentence.

with retry data, the accuracy easily goes up in Figure 4(b) (and this holds even for beam=1).[20]

Similarly, "retry upon regret" lets the model re-generate another solution step if error is detected, and this for instance gives only accuracy gain 78% ⇒ 80% in the iGSM-med$_{pq}^{\text{op}=23}$ case (recall Figure 3(b)), while pretrain with retry data can give 78% ⇒ 95%. This is strong signal that:

> **Result 6.** *Error correction is a skill that can be **fundamentally different** from beam search or retry based on the model's randomness.*

Therefore, to truly improve the model's reasoning capability, it is crucial to modify the training data to include mistakes and their corrections.

## 5 Result 7: Finetune with Retry Data

In this section, we consider a model *pretrained* using only error-free math data but *finetuned* with the retry data from Section 4. This simulates a real-world scenario where one is given an open-source pretrained model and wants to finetune it for better reasoning/math accuracies. Our goal is to determine if this approach works as well as pretraining directly on the retry data.

We focus on parameter-efficient fine-tuning (PEFT) methods such as LoRA [10], which are widely adopted in practice. LoRA fine-tunes a small number of trainable parameters (i.e., low-rank matrix updates) on top of the original, frozen pretrained weights.

Interestingly, despite using the same high-quality retry data as in Section 4 (with or without label masking), and our best efforts in selecting LoRA fine-tune parameters (adopting various rank choices), and ensuring sufficient training (comparable training steps/samples to pretraining), we find that LoRA finetuning falls short of pretraining directly with the retry data. When the LoRA rank is small, it even underperforms compared to pretraining with error-free data, see Figure 7. From this, we conclude that:

> **Result 7** (Figure 7). *Error correction is a skill that can be very different from the original (error-free) reasoning and cannot be acquired during a LoRA finetune stage from language models pretrained only using error-free data.*

(In contrast, such a skill can be learned using full finetuning with sufficiently many retry training samples, see Figure 7; but the finetune cost of this process is no less than pretraining with retry data, and is essentially continued pretraining.[21])

One may compare this to **error detection** (see Section 3.1): error detection is an easier skill that even models pretrained from error-free data can acquire almost for free (such as via probing, not to mention LoRA finetuning). However, **error correction** cannot be easily achieved via LoRA finetuning from a language model pretrained on error-free data.

We conjecture this is because, when a mistake is made, the model needs to revise its internal computations to find alternative solutions. Such revision may not be straightforward; otherwise, simply re-generating a few times from the previous sentence (i.e., "retry upon regret") should have already achieved higher reasoning accuracies. We do not analyze such internal computations in this paper, but refer interested readers to [29], which explains language models' internal computations (via probing) on the error-free iGSM data.

---

[20]To present the cleanest result, in Figure 4(b) we present the best accuracy among beam=1 or 4; one can still observe a high accuracy boost for beam=1 when pretrained with retry data.

[21]Specifically, from a model pretrained with $T$ tokens of error-free data, we additionally full finetune it with $T$ tokens of retry data and present this as "continued pretraining" in Figure 7. In Figure 10 of Appendix D.3, we additionally show that such full finetuning is no better than directly pretraining with $2T$ tokens from retry data.

| | iGSM-med_pq | | iGSM-med_qp | | iGSM-hard_pq | | iGSM-hard_qp | | row |
| --- | --- | --- | --- | --- | --- | --- | --- | --- | --- |
| | in-dist | out-of-dist (OOD) | in-dist | out-of-dist (OOD) | in-dist | out-of-dist (OOD) | in-dist | out-of-dist (OOD) | avg |
| original | 99.9 99.1 | 92.0 88.4 84.5 77.7 91.6 | 100 99.3 | 92.4 89.9 84.8 78.3 91.4 | 100 99.4 | 94.9 93.7 91.4 88.4 84.0 91.4 | 100 99.4 | 94.5 93.3 91.8 88.3 85.3 89.4 | 91.8 |
| retryrate0.2 | retry (finetune, lora qv4e8) | 99.9 98.1 | 84.0 77.5 71.8 63.3 92.7 | 99.0 98.0 | 81.6 76.0 68.4 59.4 90.3 | 99.9 98.9 | 89.2 86.1 82.1 77.3 71.9 94.5 | 99.8 98.4 | 90.1 85.6 81.6 77.5 70.3 92.1 | 85.2 |
| retryrate0.2 | retry (finetune, lora qv4e8) | with mask | 100 99.0 | 91.2 88.2 82.4 74.3 93.5 | 99.8 99.3 | 90.6 86.0 80.5 73.0 91.8 | 100 99.3 | 94.2 92.5 89.5 85.3 81.7 92.8 | 100 99.4 | 94.0 91.9 89.4 84.8 79.8 92.3 | 90.5 |
| retryrate0.2 | retry (finetune, lora qv8e16) | 100 98.6 | 86.4 82.4 76.0 68.9 93.2 | 99.6 98.4 | 84.9 79.4 72.3 63.5 91.5 | 100 99.1 | 90.9 88.0 84.3 79.9 75.5 95.2 | 99.8 98.6 | 89.9 87.4 84.1 79.9 75.0 90.5 | 87.1 |
| retryrate0.2 | retry (finetune, lora qv8e16) | with mask | 100 99.2 | 92.0 88.7 83.7 76.7 93.2 | 99.8 99.4 | 92.2 88.0 82.2 75.3 92.1 | 100 99.4 | 95.1 93.6 91.6 87.3 83.4 92.7 | 99.9 99.4 | 95.1 93.4 91.5 87.9 84.3 92.2 | 91.6 |
| retryrate0.2 | retry (finetune, lora qv16e32) | 100 98.8 | 88.3 84.4 78.4 72.0 93.8 | 99.8 98.7 | 87.8 82.8 76.3 68.1 91.3 | 100 99.2 | 94.0 92.3 89.9 85.9 83.5 96.0 | 100 99.1 | 93.0 90.6 87.7 83.6 79.5 92.4 | 89.6 |
| retryrate0.2 | retry (finetune, lora qv16e32) | with mask | 100 99.3 | 91.8 88.7 83.6 75.9 93.8 | 99.9 99.3 | 92.0 88.3 83.0 76.4 92.0 | 99.9 99.4 | 95.7 94.4 92.5 88.8 85.4 93.1 | 100 99.5 | 95.4 93.8 92.3 88.8 85.5 92.1 | 92.0 |
| retryrate0.2 | retry (finetune, lora qv32e64) | 100 98.7 | 89.8 86.2 81.1 74.7 93.0 | 99.8 99.0 | 87.9 82.4 76.5 69.9 91.5 | 100 99.4 | 95.1 92.9 90.3 88.4 84.8 96.2 | 100 99.1 | 94.3 92.6 90.3 86.1 82.2 92.4 | 90.5 |
| retryrate0.2 | retry (finetune, lora qv32e64) | with mask | 100 99.2 | 92.4 89.4 84.2 77.8 93.4 | 99.9 99.5 | 92.9 88.9 83.9 77.2 92.2 | 100 99.5 | 95.6 94.1 92.2 88.7 85.3 93.6 | 99.8 99.5 | 96.3 94.3 93.1 89.3 86.1 92.4 | 92.3 |
| retryrate0.2 | retry (finetune, lora qv64e128) | 99.9 99.1 | 90.3 86.1 81.1 76.5 92.7 | 99.9 99.1 | 88.9 84.5 78.4 71.4 92.0 | 100 99.6 | 95.3 93.6 91.5 88.4 85.1 96.5 | 99.8 99.3 | 95.3 93.1 91.3 87.7 84.8 92.3 | 91.1 |
| retryrate0.2 | retry (finetune, lora qv64e128) | with mask | 100 99.4 | 92.6 89.8 85.4 79.3 93.2 | 99.9 99.5 | 92.8 88.9 83.8 77.5 92.4 | 100 99.5 | 95.7 94.4 92.1 88.0 84.9 94.0 | 100 99.7 | 96.5 94.8 93.3 90.2 87.3 92.4 | 92.6 |
| retryrate0.2 | retry (finetune, lora qv128e256) | 99.9 99.0 | 90.8 86.9 82.2 76.5 93.9 | 99.8 99.2 | 90.6 86.9 81.9 75.4 92.0 | 100 99.6 | 96.1 94.7 92.7 90.2 87.8 96.3 | 99.8 99.4 | 95.9 94.3 92.2 89.5 86.3 92.0 | 92.1 |
| retryrate0.2 | retry (finetune, lora qv128e256) | with mask | 100 99.4 | 92.9 90.0 84.8 79.0 93.9 | 99.9 99.6 | 92.7 88.9 84.3 78.3 92.7 | 100 99.6 | 96.1 94.8 93.2 89.9 86.8 94.2 | 99.8 99.7 | 96.3 95.1 93.8 90.6 88.0 92.3 | 92.9 |
| retryrate0.2 | retry (finetune, lora qv256e512) | 100 99.1 | 92.0 87.9 84.1 78.8 94.4 | 99.9 99.2 | 90.1 85.5 80.7 74.1 92.1 | 100 99.4 | 95.5 93.1 91.4 88.8 85.3 95.8 | 99.9 99.5 | 95.6 94.0 92.3 88.8 86.0 91.9 | 91.8 |
| retryrate0.2 | retry (finetune, lora qv256e512) | with mask | 100 99.3 | 92.5 90.1 85.5 79.1 94.4 | 99.9 99.6 | 92.9 88.6 84.4 77.3 92.5 | 99.9 99.5 | 96.0 94.6 92.4 88.8 85.5 94.2 | 99.9 99.6 | 96.6 94.9 93.6 90.4 87.3 92.6 | 92.7 |
| retryrate0.2 | retry (continued pretrain) | 100 99.9 | 97.4 95.6 92.8 88.3 95.9 | 100 100 | 97.9 95.0 92.5 88.6 93.7 | 100 100 | 99.2 98.5 97.9 97.0 95.6 97.5 | 100 100 | 99.1 98.7 98.2 97.5 96.3 94.2 | 96.9 |
| retryrate0.2 | retry (continued pretrain) | with mask | 100 99.9 | 97.0 94.7 91.7 87.7 95.7 | 100 99.9 | 97.7 94.9 92.6 88.2 93.5 | 100 100 | 99.4 98.8 97.9 97.2 95.7 98.0 | 100 100 | 99.3 98.8 98.6 98.0 96.5 94.6 | 96.9 |
| retryrate0.2 | retry (pretrain) | 100 99.9 | 97.9 96.8 95.2 91.6 96.0 | 100 99.8 | 96.7 94.5 92.0 88.3 93.4 | 100 99.8 | 97.9 96.6 95.8 93.7 91.7 95.6 | 100 99.9 | 98.4 97.0 96.4 95.0 92.1 92.9 | 96.2 |
| retryrate0.2 | retry (pretrain) | with mask | 100 99.9 | 97.2 95.5 93.0 87.8 95.8 | 100 99.9 | 97.4 96.2 94.4 90.9 93.7 | 100 99.9 | 98.6 97.1 96.5 95.0 92.9 95.3 | 100 99.9 | 98.5 97.5 96.8 95.3 94.0 91.3 | 96.3 |

op≤15  op=15  op=20  op=21  op=22  op=23  op≤15  op=15  op=20  op=21  op=22  op=23  op≤21  op=21  op=28  op=29  op=30  op=31  op=32  op≤21  op=21  op=28  op=29  op=30  op=31  op=32  
op=20 (reask)   op=20 (reask)   op=28 (reask)   op=28 (reask)

Figure 7: Pretrain vs. finetune on retry data. Details and more results (for retry_rate $\neq 0.2$) are in Appendix D.3.

---

The **original** and **retry (pretrain) rows** are pretraining with error-free / retry data (same as Figure 4(b)).

---

The **retry (finetune, lora) rows** correspond to LoRA finetuning using the retry data on a model pretrained with error-free data. **qv** stands for the LoRA rank on the query/value matrices, and **e** stands for the LoRA rank on the embedding matrix. **Observation:** No significant improvement over the original model; for small LoRA ranks, finetuning even hurts and label masking becomes important.

---

The **retry (continued pretrain) rows** correspond to full finetuning using the retry data on a model pretrained with error-free data. **Note: this is for illustration only.** Our full finetuning (+ pretrain) uses twice the training tokens compared to **retry (pretrain)**, so it is not surprising that accuracies can be higher. We compare this to directly pretraining with twice the retry data tokens in Figure 10.

---

**Conclusion:** Error correction is a skill very different from the original (error-free) reasoning and may not be acquired during a LoRA finetune stage, even with a sufficient number of finetune (retry) samples.

## 6 Result 8: Pretrain with Fake Mistakes

While it is possible to prepare "perfect" retry data on synthetic iGSM datasets, obtaining math data with mistakes and corrections can be challenging in real life. For this reason, we explore the possibility of using more realistic approaches to augment math problems (for which we only have correct solutions) with **"fake" mistakes and retries**. Ideally, this process should not require any semantic parsing or understanding of the problem and the solution.

We explore two approaches and compare them with the perfect retry data. We still use retry_rate, but instead of selecting a wrong parameter to retry, we simplify the process:

- In the first approach, we randomly select a sentence that appears *later* in the step-by-step solution to retry. For instance, in the example of Figure 2:

  **(Solution - retry_weak)** Define Dance Studio's School Daypack as p; so p = 17. Define Film Studio's Messenger Backpack as W; so W = 13. **Define ◇◇◇ as [BACK].** Define Central High's Film Studio as B; so B = p + W = 17 + 13 = 7. Define Film Studio's School Daypack as g; R = W + B = 13 + 7 = 20; so g = 12 + R = 12 + 20 = 9. Define Film Studio's Backpack as w; so w = g + W = 9 + 13 = 22. Define Central High's Backpack as c; so c = B * w = 7 * 22 = 16. *Answer: 16.*

  (6.1)

  At the ◇◇◇ position, we randomly select one of the three underlined parameters and add it with [BACK] for retry.[22] At the end of each sentence, we add at least one such retry parameter with retry_rate probability, at least two with (retry_rate)$^2$ probability, and so on.

  This approach introduces *fake* mistakes because a parameter in a later sentence might still be

---

[22] Alternatively, one can also insert the entire sentence (as opposed to just the parameter name) for retry.

| | iGSM-med_pq | | | | | | | iGSM-med_qp | | | | | | | iGSM-hard_pq | | | | | | | iGSM-hard_qp | | | | | | | row avg |
| | in-dist | out-of-dist (OOD) | | | | | | in-dist | out-of-dist (OOD) | | | | | | in-dist | out-of-dist (OOD) | | | | | | in-dist | out-of-dist (OOD) | | | | | | |
| | op ≤ 15 | op=15 | op=20 | op=21 | op=22 | op=23 | op=20 (reask) | op ≤ 15 | op=15 | op=20 | op=21 | op=22 | op=23 | op=20 (reask) | op ≤ 21 | op=21 | op=28 | op=29 | op=30 | op=31 | op=32 | op=28 (reask) | op ≤ 21 | op=21 | op=28 | op=29 | op=30 | op=31 | op=32 | op=28 (reask) | |
|---|---|---|---|---|---|---|---|---|---|---|---|---|---|---|---|---|---|---|---|---|---|---|---|---|---|---|---|---|---|---|---|
| original | 99.9 | 99.1 | 92.0 | 88.4 | 84.5 | 77.7 | 91.6 | 100 | 99.3 | 92.4 | 89.9 | 84.8 | 78.3 | 91.4 | 100 | 99.4 | 94.9 | 93.7 | 91.4 | 88.4 | 84.0 | 91.4 | 100 | 99.4 | 94.5 | 93.3 | 91.8 | 88.3 | 85.3 | 89.4 | 91.8 |
| retry_miss | retryrate0.05 | 100 | 99.3 | 92.1 | 88.3 | 82.6 | 76.3 | 93.1 | 99.9 | 99.3 | 91.1 | 85.0 | 78.0 | 70.7 | 91.8 | 99.9 | 99.3 | 94.8 | 91.7 | 89.5 | 86.6 | 82.3 | 90.3 | 100 | 99.3 | 95.5 | 93.5 | 92.5 | 89.4 | 86.4 | 91.6 | 91.0 |
| retry_miss | retryrate0.1 | 100 | 99.6 | 94.3 | 91.7 | 88.1 | 81.6 | 91.4 | 99.9 | 99.4 | 93.3 | 88.9 | 85.5 | 79.4 | 91.5 | 100 | 99.8 | 96.0 | 95.2 | 92.3 | 90.2 | 87.3 | 91.7 | 100 | 99.2 | 95.5 | 92.3 | 91.0 | 87.6 | 83.8 | 90.5 | 92.6 |
| retry_miss | retryrate0.2 | 100 | 99.7 | 96.1 | 94.1 | 91.2 | 87.1 | 93.5 | 100 | 99.5 | 93.3 | 89.1 | 86.0 | 79.4 | 91.7 | 100 | 99.7 | 96.2 | 93.6 | 92.1 | 89.0 | 84.3 | 92.2 | 100 | 99.7 | 97.4 | 95.4 | 94.2 | 92.0 | 89.5 | 90.9 | 93.5 |
| retry_miss | retryrate0.5 | 100 | 99.3 | 93.2 | 91.7 | 88.7 | 83.2 | 93.5 | 100 | 98.9 | 91.2 | 86.7 | 82.8 | 76.3 | 91.8 | 100 | 99.0 | 90.3 | 87.4 | 85.2 | 82.0 | 79.9 | 91.4 | 100 | 98.7 | 85.5 | 80.5 | 78.7 | 73.7 | 68.7 | 89.2 | 88.9 |
| retry_weak | retryrate0.05 | 100 | 99.7 | 95.8 | 93.5 | 90.7 | 86.4 | 95.9 | 99.9 | 99.5 | 94.4 | 90.5 | 86.2 | 79.5 | 91.5 | 100 | 99.7 | 97.5 | 96.0 | 94.8 | 92.7 | 89.6 | 95.6 | 100 | 99.7 | 97.2 | 94.8 | 93.2 | 90.2 | 87.5 | 91.5 | 94.1 |
| retry_weak | retryrate0.05 (with mask) | 100 | 99.0 | 93.0 | 90.0 | 84.5 | 78.1 | 91.6 | 99.8 | 99.4 | 93.0 | 89.2 | 84.3 | 75.6 | 90.9 | 100 | 99.8 | 96.9 | 95.6 | 94.4 | 92.0 | 88.5 | 94.8 | 100 | 99.7 | 97.3 | 95.2 | 93.5 | 92.2 | 89.5 | 91.9 | 93.0 |
| retry_weak | retryrate0.1 | 100 | 99.8 | 97.2 | 96.3 | 94.0 | 90.1 | 95.4 | 100 | 99.8 | 96.6 | 94.3 | 91.9 | 88.3 | 92.4 | 100 | 99.9 | 98.1 | 97.0 | 96.2 | 94.0 | 92.2 | 93.2 | 100 | 99.8 | 97.4 | 95.4 | 93.9 | 91.4 | 89.7 | 91.2 | 95.5 |
| retry_weak | retryrate0.1 (with mask) | 100 | 99.8 | 97.1 | 96.1 | 94.2 | 90.1 | 95.4 | 99.9 | 99.8 | 96.1 | 93.8 | 91.5 | 86.4 | 93.2 | 100 | 99.7 | 97.4 | 96.1 | 94.6 | 93.5 | 90.6 | 91.5 | 100 | 99.8 | 97.7 | 96.3 | 95.4 | 93.1 | 91.4 | 90.1 | 95.3 |
| retry_weak | retryrate0.2 | 100 | 99.8 | 97.5 | 96.7 | 94.6 | 91.0 | 94.9 | 99.9 | 99.8 | 94.8 | 92.5 | 88.0 | 82.7 | 92.1 | 99.9 | 99.7 | 97.1 | 94.7 | 93.8 | 91.2 | 88.8 | 92.2 | 100 | 99.8 | 98.4 | 97.6 | 96.3 | 95.3 | 94.1 | 90.9 | 95.1 |
| retry_weak | retryrate0.2 (with mask) | 100 | 99.8 | 97.5 | 96.3 | 93.9 | 90.7 | 94.3 | 100 | 99.8 | 95.1 | 92.3 | 87.5 | 82.8 | 91.0 | 100 | 99.8 | 97.1 | 94.9 | 93.9 | 91.7 | 88.5 | 93.2 | 99.9 | 99.8 | 97.9 | 97.1 | 95.9 | 94.7 | 92.8 | 91.5 | 95.0 |
| retry_weak | retryrate0.5 | 100 | 98.2 | 80.0 | 73.5 | 66.0 | 61.1 | 89.9 | 100 | 96.3 | 71.9 | 67.6 | 59.9 | 54.7 | 88.2 | 99.6 | 89.7 | 66.7 | 59.2 | 56.0 | 52.7 | 46.0 | 90.1 | 99.5 | 90.3 | 69.2 | 62.9 | 59.2 | 54.3 | 50.9 | 85.7 | 74.6 |
| retry_weak | retryrate0.5 (with mask) | 98.9 | 85.7 | 64.5 | 56.2 | 51.4 | 43.9 | 86.3 | 99.9 | 97.9 | 79.6 | 71.7 | 64.4 | 56.2 | 87.6 | 99.6 | 89.0 | 69.4 | 63.5 | 60.0 | 56.5 | 51.6 | 87.7 | 99.4 | 88.4 | 67.7 | 61.7 | 58.8 | 55.3 | 51.3 | 87.7 | 73.1 |

Figure 8: Accuracies on models pretrained in retry_weak or retry_miss data vs. the original no-mistake data.

> **Observations.** The retry_weak format of pretrain data significantly improves accuracy (especially on the hardest op = 23/32 tasks) and is easy to obtain in practice. There is no need to use retry_miss data, which is slightly harder to acquire. Experiment details are in Appendix D.1.

computable at the current position. However, it is very easy to implement: one simply needs to select a future sentence and append it with a [BACK] symbol. Intuitively, it encourages the model *not to skip steps*. We call this **retry_weak** data.

- In the second approach, we randomly select a parameter that appears in the problem statement and has not yet appeared in the solution. For instance in (6.1), at location ◇◇◇, one can add not only one of the underlined parameters but also any other parameter from the problem that hasn't appeared before (such as Riverview High's Film Studio). We call this **retry_miss** data. This type of data is harder to obtain than retry_weak but easier than the perfect retry data. Intuitively, it encourages the model not only "not to skip steps" but also not to compute unnecessary parameters.

Our results are in Figure 8 (cf. Figure 4(b) for the perfect retry data); to summarize:

> **Result 8.** *The realistic, simple-to-obtain* retry_weak *data significantly improve the model's accuracy; yet, the slightly more complex* retry_miss *data does not improve accuracy by much.*

(Additionally, Figure 11(a) in Appendix C shows that using retry_weak/retry_miss data, the model has a higher retry rate at inference time compared to Figure 5(a); Figure 11(b) shows in both cases, the model still learns to find shortest solutions, similar to Figure 6 on perfect retry data.)

Please note, while our experiments are on synthetic data, we aim to use controlled experiments to predict what could be the important data changes that can help improve LLMs' reasoning capabilities in real life. While future LLMs may not be trained directly on such retry_weak data, our results suggest that it can be beneficial to, for instance, use auxiliary models to rewrite math data to include fake mistakes of this type.

## 7 Conclusion

In this paper, we investigate whether language models can benefit from pretraining on data containing mistakes, followed by immediate error correction. Using a fully controllable synthetic setting, we demonstrate that models trained on such data outperform those trained on the same amount of error-free data.

In addition to the accuracy gain, Section 4 shows that using retry data is very safe: the model rarely makes mistakes even after pretraining with high error-rate retry data, and it is unnecessary to change the training process (simply autoregressive, no need to label-mask the errors). Retry data teaches models how to correct errors if needed, rather than encouraging mistakes.

It is important to note that such *error correction skill* does not come easily. A model pretrained with only error-free data cannot use (1) beam search or (2) retry based on error detection ("retry upon regret") to achieve comparable performance, see Section 3, unless the error detection is nearly perfect. This error correction skill is also very different from the original error-free reasoning and thus cannot be learned during parameter-efficient fine-tuning (PEFT) such as LoRA, see Section 5. This implies the necessity of adding retry data to the pretrain data for language models to truly learn the capability to correct errors.

While grade-school level math problems have many other difficulties (including arithmetic or common sense), following [29], we have focused on the (logic-following) reasoning aspect, which is one of the weakest aspects of GPT-4.[23]

While it is unlikely that iGSM retry data will be directly used for pretraining future commercial-level LLMs, this paper aims to find guiding principles for necessary ingredients. We strongly discourage using fine-tuning to teach a model to correct errors (or using beam search, or letting the model regenerate upon encountering a mistake) as these are not effective. We advocate for adding mistakes and corrections at the pretrain level. While commercial LLMs use synthetic data [16, 25] and future LLMs are rumored to use $Q^\star$, it remains a question of how to prepare such synthetic data for the model to best learn error correction. Our Section 6 suggests that it is critical to teach a model not to skip steps. This can be done either through naively creating retry_weak data like ours or using more advanced prompting methods to encourage an auxiliary model to rewrite math data into such a format. We cannot explore such follow-up directions due to GPU resource limitations.

Finally, Part 2 of this work series focuses on how language models solve grade-school math level reasoning problems (including Part 2.1 [29]). We also cover how language models learn language structures in Part 1 [2] and learn world knowledge in Part 3 [3–5].

---

[23]In contrast, arithmetic such as 10-digit multiplications can be (and perhaps should be) handled by calculators to save the model's capacity for other skills.

# Appendix

## A  A Harder Example with Retry

**(Problem)** The number of each Jungle Jim's International Market's Cheese equals the sum of each Parmesan Cheese's Pear and each The Fresh Market's Ice Cream. The number of each Ice Cream's Pineapple equals 2 more than each Goat Cheese's Grape. The number of each New Seasons Market's Goat Cheese equals the sum of each Residential College District's Jungle Jim's International Market, each Jungle Jim's International Market's Parmesan Cheese and each Residential College District's Supermarket. The number of each Arts Campus's New Seasons Market equals each Cheese's Pineapple. The number of each Goat Cheese's Banana equals each Vocational School District's Product. The number of each Residential College District's Jungle Jim's International Market equals 5 more than each Ice Cream's Grape. The number of each Parmesan Cheese's Pineapple equals each Parmesan Cheese's Pear. The number of each Residential College District's The Fresh Market equals each Arts Campus's Trader Joe's. The number of each Arts Campus's Trader Joe's equals each Parmesan Cheese's Ingredient. The number of each Goat Cheese's Grape equals 0. The number of each The Fresh Market's Ice Cream equals 13 more than the difference of each Residential College District's The Fresh Market and each Parmesan Cheese's Grape. The number of each Goat Cheese's Pineapple equals each New Seasons Market's Product. The number of each Vocational School District's The Fresh Market equals the sum of each Trader Joe's's Cheese and each The Fresh Market's Cheese. The number of each Trader Joe's's Cheese equals 6. The number of each The Fresh Market's Cheese equals 3. The number of each Jungle Jim's International Market's Ice Cream equals the difference of each Ice Cream's Banana and each Goat Cheese's Grape. The number of each Jungle Jim's International Market's Parmesan Cheese equals each Ice Cream's Pineapple. The number of each Parmesan Cheese's Pear equals the difference of each Goat Cheese's Grape and each Ice Cream's Grape. The number of each Parmesan Cheese's Grape equals 12 times as much as each Residential College District's Jungle Jim's International Market. The number of each The Fresh Market's Parmesan Cheese equals each The Fresh Market's Cheese. The number of each Ice Cream's Banana equals the sum of each Parmesan Cheese's Pineapple and each Ice Cream's Pineapple. The number of each School District's Jungle Jim's International Market equals each The Fresh Market's Ice Cream. The number of each Cheese's Pineapple equals 20 more than the sum of each Trader Joe's's Cheese and each The Fresh Market's Cheese. The number of each Trader Joe's's Parmesan Cheese equals 16. The number of each Ice Cream's Pear equals 8. The number of each Ice Cream's Grape equals each Goat Cheese's Grape. *How many Product does School District have?*

**(Solution)** Define Goat Cheese's Grape as u; so u = 0. Define Ice Cream's Grape as x; so x = u = 0. Define Residential College District's Jungle Jim's International Market as N; so N = 5 + x = 5 + 0 = 5. Define Parmesan Cheese's Pear as G; so G = u - x = 0 - 0 = 0. Define Parmesan Cheese's Grape as f; so f = 12 * N = 12 * 5 = 14. Define Parmesan Cheese's Pineapple as C; so C = G = 0. Define Parmesan Cheese's Ingredient as Z; e = f + C = 14 + 0 = 14; so Z = e + G = 14 + 0 = 14. Define Arts Campus's Trader Joe's as q; so q = Z = 14. Define Residential College District's The Fresh Market as j; so j = q = 14. Define Ice Cream's Pineapple as X; so X = 2 + u = 2 + 0 = 2. Define Ice Cream's Banana as K; so K = C + X = 0 + 2 = 2. Define The Fresh Market's Ice Cream as P; i = j - f = 14 - 14 = 0; so P = 13 + i = 13 + 0 = 13. Define Jungle Jim's International Market's Ice Cream as R; so R = K - u = 2 - 0 = 2. Define School District's Jungle Jim's International Market as V; so V = P = 13. Define Jungle Jim's International Market's Cheese as v; so v = G + P = 0 + 13 = 13. Define Jungle Jim's International Market's Parmesan Cheese as S; so S = X = 2. Define Jungle Jim's International Market's Product as y; U = S + R = 2 + 2 = 4; so y = U + v = 4 + 13 = 17. Define School District's Product as J; so J = V * y = 13 * 17 = 14. *Answer: 14.*

**(Solution with** retry_rate = 0.2**)** ~~Define Arts Campus's Ingredient as~~ [BACK]. ~~Define Vocational School District's Supermarket as~~ [BACK]. ~~Define Jungle Jim's International Market's Cheese as~~ [BACK]. Define Goat Cheese's Grape as u; so u = 0. Define Ice Cream's Grape as x; so x = u = 0. Define Residential College District's Jungle Jim's International Market as N; so N = 5 + x = 5 + 0 = 5. ~~Define New Seasons Market's Product as~~ [BACK]. Define Parmesan Cheese's Pear as G; so G = u - x = 0 - 0 = 0. Define Parmesan Cheese's Grape as f; so f = 12 * N = 12 * 5 = 14. Define Parmesan Cheese's Pineapple as C; so C = G = 0. Define Parmesan Cheese's Ingredient as Z; e = f + C = 14 + 0 = 14; so Z = e + G = 14 + 0 = 14. Define Arts Campus's Trader Joe's as q; so q = Z = 14. Define Residential College District's The Fresh Market as j; so j = q = 14. ~~Define Jungle Jim's International Market's Product as~~ [BACK]. Define Ice Cream's Pineapple as X; so X = 2 + u = 2 + 0 = 2. Define Ice Cream's Banana as K; so K = C + X = 0 + 2 = 2. Define The Fresh Market's Ice Cream as P; i = j - f = 14 - 14 = 0; so P = 13 + i = 13 + 0 = 13. Define Jungle Jim's International Market's Ice Cream as R; so R = K - u = 2 - 0 = 2. ~~Define Vocational School District's Supermarket as~~ [BACK]. Define School District's Jungle Jim's International Market as V; so V = P = 13. ~~Define New Seasons Market's Ingredient as~~ [BACK]. Define Jungle Jim's International Market's Cheese as v; so v = G + P = 0 + 13 = 13. Define Jungle Jim's International Market's Parmesan Cheese as S; so S = X = 2. Define Jungle Jim's International Market's Product as y; U = S + R = 2 + 2 = 4; so y = U + v = 4 + 13 = 17. Define School District's Product as J; so J = V * y = 13 * 17 = 14.

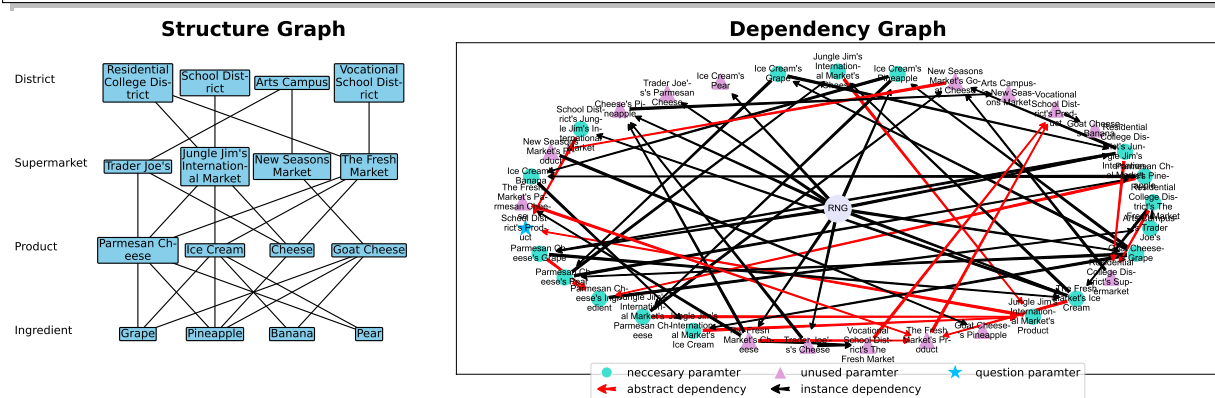

Figure 9: A harder example with op = 21 in iGSM-hard$_{pq}$ used for training. We also provide a retry example here with retry_rate = 0.2.

# B  More Experiments for Finetune (Result 7)

We have included the finetune results only for retry_rate $= 0.2$ in Figure 7 in Section 5 (Result 7). In Figure 10 below, we include the additional results for retry_rate $\neq 0.2$.

Figure 10: Pretrain vs. finetune on retry data. This figure complements Figure 7 by including additional results for retry_rate $\neq 0.2$. All conclusions under Figure 7 remain the same.

> **Additional observation 1:** although rarely, **retry (finetune, lora)** can outperform the original model especially when retry_rate $= 0.5$, *and* label masking is on, *and* a very high-rank LoRA finetune is used.
>
> **Additional observation 2:** although **retry (continued pretrain)** improves upon **retry (pretrain)**, it uses twice the number of training tokens ($T$ tokens from retry plus $T$ tokens from error-free data); if comparing this against **retry (pretrain, double-time)** which is to directly pretrain using $2T$ tokens from retry data, there is no significant accuracy improvement.

# C  More Experiments for Fake Retry Data (Result 8)

We have included the accuracy results for the retry_weak or retry_miss data in Figure 8 in Section 6 (Result 8). Below in Figure 11, we additionally showcase the number of retries and that the models can still generate shortest solutions most of the time.

(a) Model's average number of retries per test problem (among generated correct solutions); this figure complements Figure 5(a) by including also retry_weak and retry_miss.

(b) Model's average number of unnecessary operations or parameters per test problem (among correct solutions); this figure complements Figure 6 by including also retry_weak and retry_miss.

Figure 11: Average number of retries and unnecessary operations per generated (correct) solution. Details see Appendix D.1.

> **Observations.** When pretrained with retry_weak or retry_miss data, the model tends to retry more frequently than the perfect retry data, especially for retry_rate $\geq 0.1$ (top table); yet the model does not tend to compute more unnecessary parameters than the perfect retry data (bottom table).

# D   Experiment Details and Parameters

**Model.**   We use the GPT2 architecture [20], replacing its absolute positional embedding with modern rotary positional embedding [7, 24], still referred to as GPT2 for short. We also played with the Llama architecture (especially with gated MLP layers) and did not see any benefit of using it. This GPT2 performs comparably to Llama/Mistral at least for knowledge tasks [5].

Let GPT2-$\ell$-$h$ denote an $\ell$-layer, $h$-head, $64h$-dim GPT2 model. We primarily use GPT2-12-12 (a.k.a. GPT2-small) throughout this paper. We use the default GPT2Tokenizer, and a context length of 768/1024 for language model pretraining on iGSM-med/iGSM-hard and a context length of 2048 for evaluation.

**Data Size.**   For both pretraining and finetuning, we did not limit the amount of training data; we generated new data on-the-fly. We do not explore sample complexity in this paper, such as the number of math problems needed to achieve a certain level of accuracy, as it would complicate the main message of this paper.

## D.1   Pretrain Experiment Details

**Pretrain parameters.**   Throughout this paper, when we pretrained a model (except for "pretrain double-time"), we followed the same pretrain parameter choices of [29]. That is, we used the AdamW optimizer with mixed-precision fp16, $\beta = (0.9, 0.98)$, cosine learning rate decay (down to 0.01x in the end + 1000 steps of linear ramp-up). We used a mixture of V100/A100 GPUs, but the GPU specifications are not relevant here.[24] For all the experiments (with original data, or retry data, or retry_weak, or retry_miss):

- For pretraining on the iGSM-med datasets, we used a learning rate of 0.002, weight decay of 0.05, batch size of 512, context length of 768, and trained for $100,000$ steps.

- For pretraining on the iGSM-hard datasets, we used a learning rate of 0.002, weight decay of 0.03, batch size of 256, context length of 1024, and trained for $200,000$ steps.

Our pretrain data is constructed by randomly generating math problems (and solutions), concatenating them together, and truncating them (in the right) to fit within the 768 or 1024-sized context window. If a problem (with solution) is longer than the context window size, we discard it.

**Test-time parameters.**   When evaluating on test data, we use context length 2048 for both iGSM-med and iGSM-hard.[25] We use either beam=1 and dosample=false (greedy) or beam=4 and dosample=true (beam-search multinomial sampling) to present test accuracies. (Except for the original no-retry training, we also tried beam=16/32 with dosample=true.) We discover it is better to keep dosample=false while beam=1 and dosample=true while beam > 1.

Our accuracies *are not simply* from comparing the answer integers (between 0 and 22); instead we have written a parser to make sure the model's intermediate solution steps are fully-correct.

**Accuracy statistics.**   In each of our accuracy result, such as each cell in Figure 3(b), Figure 4(b), Figure 8, we average the model's evaluation over 4096 math problems of that type.

---

[24]A 128-GPU job with batch size 1 each would be identical to a 32-GPU job with batch size 4 each.

[25]During evaluation, we discard problems (with error-free, ground-truth solutions) longer than 768 tokens for iGSM-med (or 1024 for iGSM-hard), but allow the generation process to use up to 2048 tokens. This follows [29] to ensure a direct comparison. Notably, we *do not* discard problems based on their token length after including errors and corrections. Adding errors lengthens the solutions, and discarding such data would bias the test distribution towards simpler problems. Thus, when comparing models pretrained with error-free vs retry data in Figure 4(b), the models were *evaluated using the same test data distribution*.

In Figure 4(b) and Figure 8, since we care about the (relatively small) accuracy differences across models, we pretrain using two different random seeds, and evaluate with both beam=1/4; we then present the best accuracies in each cell with respect to the 2 seeds and 2 beam choices.

For the retry count statistics in Figure 5(a), Figure 5(b), Figure 11(a) and the unnecessary parameter count statistics in Figure 6 and Figure 11(b), we also tested each model with 4096 math problems in each case. We then presented the statistics among the model's correct solutions or wrong solutions in these figures.

**Pretrain (double-time) parameters.** In Figure 10 of this appendix, we have also included experiments with pretraining for twice the number of retry data tokens. In that experiment, we have followed the same pretraining parameters, except that:

- For pretraining (double-time) on the iGSM-med datasets, we decreased weight decay to 0.03, and trained for $200,000$ steps (twice than before).
- For pretraining (double-time) on the iGSM-hard datasets, we decreased weight decay to 0.02, and trained for $400,000$ steps (twice than before).

## D.2 V-Probing Experiment Details

The V-probing for can_next($A$) was introduced in Ye et al. [29]. It is a *fine-tuning* process upon the pretrained language model, with an additional linear head on the output layer, and a small rank-$r$ update on the input (embedding) layer. The pretrained model is freezed, and only this linear head and the rank-$r$ update are trainable parameters during the fine-tuning stage (for the probing task).

In this paper, our "can_next probing version1" followed the exact same parameters as [29]. That is, we use a small $r = 8$ so if probing accuracy is high, it mostly comes from the pretrained model and not the additional trainable parameters. We used learning rate 0.002 (with no ramp-up, linear decay down to 0), weight decay of 0.01, and a batch size of 256 for iGSM-med (or 128 for iGSM-hard) and trained for $100,000$ steps. One can calculate that in version1, this is 50% (or 25% for iGSM-hard) of the training tokens comparing to the pretrain process. That is a lot of finetune samples.[26]

For such reason, we also implemented a "can_next probing version2", which uses more trainable parameters but significantly fewer training samples. Specifically, we choose $r = 32$ and additionally allow for a rank-2 update on the query/value matrices in the GPT2 model (this is $768 \times 4 \times 2 \times 12$ trainable parameters for all the 12 layers). We used batch size 8, best learning rate between $\{0.001, 0.0005\}$ (same linear scheduling), weight decay 0, and finetuned for $16,000$ steps. This is only 0.25% of the training tokens comparing to the pretrain stage.

We implemented "retry upon regret" for both versions and their accuracies were presented in Figure 3(b) — once again, averaged over 4096 problems in each cell.

## D.3 Finetune Experiment Details

**LoRA finetune.** In Section 5, we applied LoRA finetuning [10] on a pretrained model using the new "error + correction" data. LoRA involves freezing the pretrained model and adding trainable low-rank updates to the weight matrices. It is recommended to apply low-rank updates to the query/value matrices [10] and the embedding matrix [4].

---

[26]The goal of [29] is to showcase the existence of *very small* rank-$r$ update, so they did not worry about the sample complexity. In this paper instead, we are interested in the practicality (e.g., to use probing to guide the model's generation) so we wish to lower down the sample complexity.

We experimented with a wide range of low-rank configurations, using rank-$r$ for the query/value matrices and rank-$2r$ for the embedding matrix, with $r \in \{4, 8, 16, 32, 64, 128, 256\}$. Notably, using $r = 256$ is almost equivalent to full finetuning, given that the hidden dimension is 768 for GPT2-12-12.

In this experiment, we used the same parameters as the pretraining (e.g., AdamW, betas, cosine lr scheduling), except that

- For LoRA finetuning on the iGSM-med datasets, we used a learning rate of 0.001, weight decay of 0.05, batch size of 256, context length of 768, and trained for 200,000 steps.
- For LoRA finetuning on the iGSM-hard datasets, we used a learning rate of 0.001, weight decay of 0.05, batch size of 128, context length of 1024, and trained for 200,000 steps.

(LoRA finetuning typically requires a smaller learning rate.) Note this is the same as the pretraining tokens for iGSM-med and half of that for iGSM-hard, which is sufficient for the training curve to plateau.[27]

Our accuracy results were presented in Figure 7, and once again in each cell we have evaluated the model over 4096 math problems. Similar to all other retry experiments (see Appendix D.1), we performed LoRA finetuning from (two) pretrained models using 2 random seeds, and presented the best accuracy among the 2 seeds and the two beam=1/4 choices.

**Full finetune = continued pretrain.** In Section 5, we also applied full finetuning. This used the same parameters as pretraining (e.g., AdamW, betas, cosine lr scheduling), except that[28]

- For full finetuning on the iGSM-med datasets, we used learning rate 0.001, weight decay 0.05, batch size of 512, context length of 768, and trained for 100,000 steps.
- For full finetuning on the iGSM-hard datasets, we used learning rate 0.001, weight decay 0.03, batch size of 256, context length of 1024, and trained for 200,000 steps.

Note that this full finetune uses the same number of training tokens comparing to pretrain in Section D.1.

Once again, our accuracy results were presented in Figure 7, and in each cell we have evaluated the model over 4096 math problems. We performed full finetuning from (two) pretrained models using 2 random seeds, and presented the best accuracy among the 2 seeds and the two beam=1/4 choices.

---

[27] We have picked the last checkpoint for evaluation (not the best); but this is okay since there is no catastrophic forgetting in this finetune process (because there are sufficiently many, freshly-sampled retry data).

[28] For iGSM-med we have also tried learning rate 0.002 and/or weight decay 0.005 and found results very similar. For iGSM-hard we also tried learning rate 0.002 and/or weight decay 0.003 and found results are very similar.